\documentclass[10pt,twocolumn,letterpaper]{article}

\usepackage{cvpr}              

\usepackage[dvipsnames]{xcolor}

\definecolor{cvprblue}{rgb}{0.21,0.49,0.74}
\usepackage[pagebackref,breaklinks,colorlinks,citecolor=cvprblue]{hyperref}

\def\paperID{*****} 
\def\confName{CVPR}
\def\confYear{2024}

\title{ACTrack: Adding Spatio-Temporal Condition for Visual Object Tracking}

\author{Yushan Han\\
Lenovo Research\\
{\tt\small hanys1@lenovo.com}
\and
Kaer Huang\\
Lenovo Research\\
{\tt\small huangke1@lenovo.com}
}

\begin{document}
\maketitle
\begin{abstract}

Efficiently modeling spatio-temporal relations of objects is a key challenge in visual object tracking (VOT). Existing methods track by appearance-based similarity or long-term relation modeling, resulting in rich temporal contexts between consecutive frames being easily overlooked. Moreover, training trackers from scratch or fine-tuning large pre-trained models needs more time and memory consumption. In this paper, we present ACTrack, a new tracking framework with additive spatio-temporal conditions. It preserves the quality and capabilities of the pre-trained Transformer backbone by freezing its parameters, and makes a trainable lightweight additive net to model spatio-temporal relations in tracking. We design an additive siamese convolutional network to ensure the integrity of spatial features and perform temporal sequence modeling to simplify the tracking pipeline. Experimental results on several benchmarks prove that ACTrack could balance training efficiency and tracking performance.
\end{abstract}    
\section{Introduction}
\label{sec:intro}
Visual object tracking (VOT) is a crucial task in the field of computer vision, which involves the process of locating and following a specific object of interest in a sequence of images or video frames. 
It has been widely used in various applications, \eg, human computer interaction, unmanned driving, robot vision, and surveillance systems. 

\begin{figure}
  \centering
  \begin{subfigure}{0.9\linewidth}
  \includegraphics[width=1\textwidth]{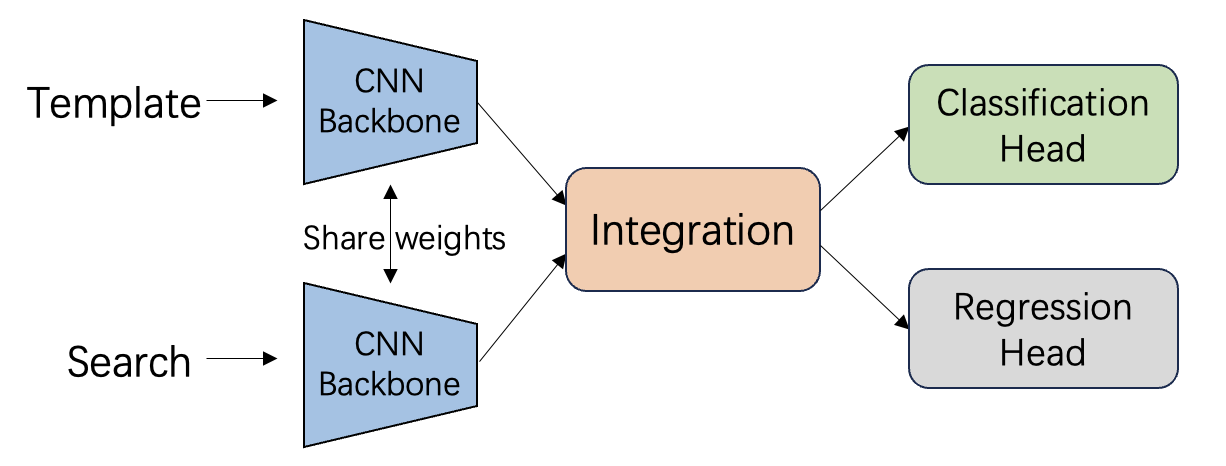}
    \caption{Trackers with siamese CNN backbone and task-specific heads.}
    \label{fig:short-a}
  \end{subfigure}
  \hfill
  \begin{subfigure}{0.9\linewidth}
\includegraphics[width=1\textwidth]{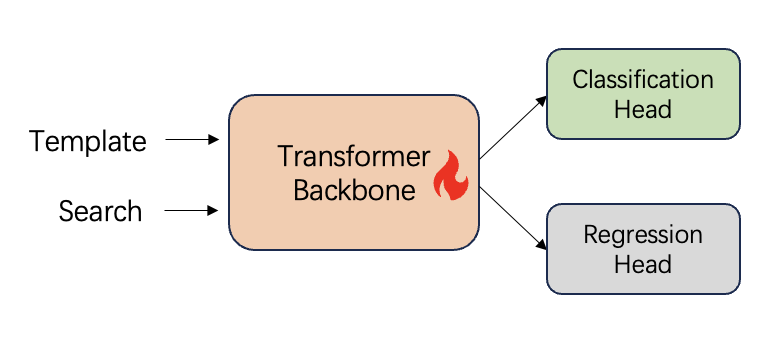}
    \caption{Trackers with Transformer backbone and task-specific heads.}
    \label{fig:short-b}
  \end{subfigure}
  
   \begin{subfigure}{0.9\linewidth}
  \includegraphics[width=1\textwidth]{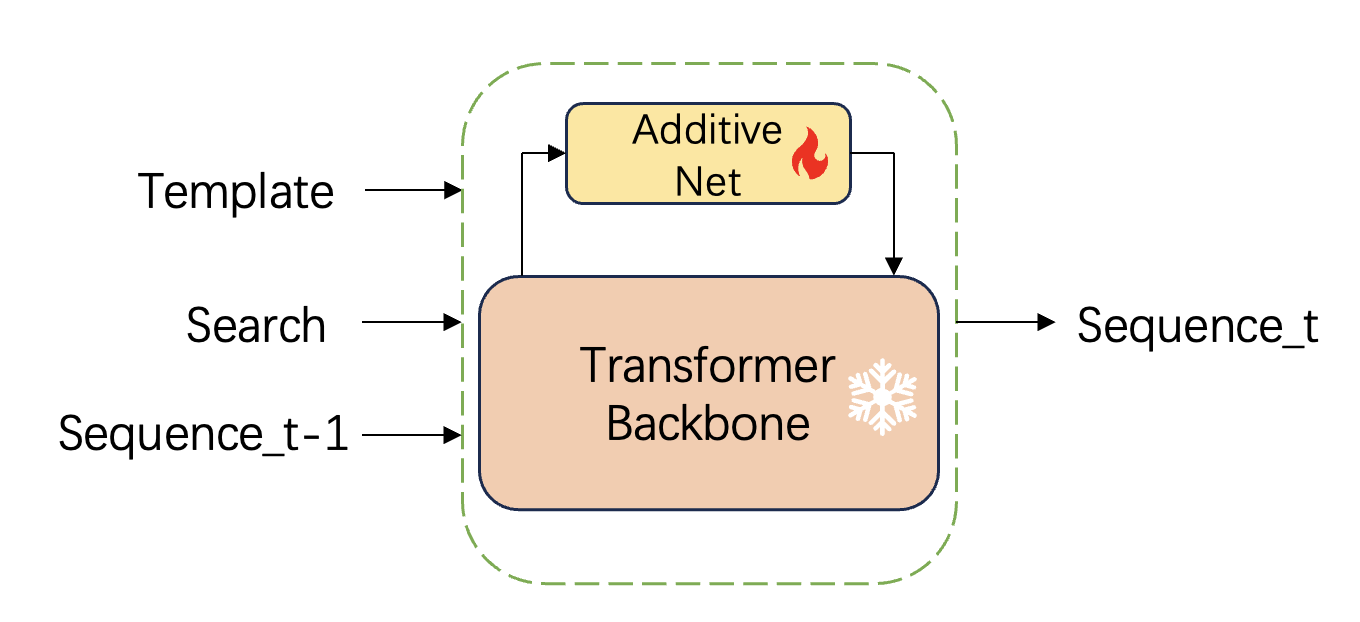}
    \caption{ACTrack: The tracker with additive spatio-temporal conditions.}
    \label{fig:short-c}
  \end{subfigure}
  \caption{Comparison of tracking frameworks. (a) The framework with three components: a CNN backbone, a feature integration module, and task-specific heads. (b) The framework with trainable Transformer backbone and task-specific heads. (c) Our ACTrack freezes the pre-trained Transformer and trains an additive lightweight conditional net to model spatio-temporal relations.}
  \label{fig:num}
\end{figure}

Most of the current prevailing trackers \cite{bertinetto2016fully,li2018high,xu2020siamfc++,chen2020siamese,guo2020siamcar,zhu2018distractor,li2019siamrpn++} follow a multi-stage pipeline, considering tracking as a problem of template-search matching. As shown in \cref{fig:short-a}, this kind of method contains several components of a siamese backbone to extract generic features, an integration module to allow information communication between tracking target and search area, and task-specific heads for object localization and scale estimation. Recently, based on its global and dynamic modeling capacity, Transformer \cite{vaswani2017attention} is introduced to the tracking field. Some works \cite{yan2021learning,chen2021transformer,huang2023multi,kristan2023first,fu2021stmtrack,gao2022aiatrack} based on CNN for consistent appearance features extraction and apply attention operations in high-level representation space. In addition, as shown in \cref{fig:short-b}, in order to simplify the tracking process, other unified frameworks \cite{chen2022backbone,cui2022mixformer,ye2022joint,wei2023autoregressive} joint feature learning and relation modeling through trainable Transformer backbone. Furthermore, several works explore the effectiveness of sequence modeling \cite{chen2023seqtrack,wei2023autoregressive} instead of explicitly performing additional classification and regression.

While the existing pipelines achieve good performance, these methods focus on long-term relation modeling and generic representation learning between template and search images. 
However, there exists rich temporal context between consecutive frames, which presents significant challenges for tracking in real-world environments due to issues such as object deformation, scale changes, occlusion, and interference from similar objects. To deal with these challenges caused by temporal variations, some techniques such as window penalty \cite{chen2021transformer,wang2021transformer,ye2022joint,zhang2020ocean}, box optimization \cite{wang2021transformer,bhat2019learning}, and template update mechanism \cite{yan2021learning,wang2021transformer,cui2022mixformer,chen2023seqtrack} have been employed. Besides, STMTrack \cite{fu2021stmtrack} and ARTrack \cite{wei2023autoregressive} adopt the design of spatio-temporal memory. However, incorporating both the modeling of the relationship of generic features and accounting for the effects of temporal variations still remains a challenge. Moreover, high-performance tracking models often have more parameters and require large-scale datasets for longer training schedules. However, training from scratch can be time-consuming, and direct fine-tuning or continued training of a large pre-trained model may cause overfitting and catastrophic forgetting \cite{hu2021lora,hayes2020remind}.


To address these challenges, in this paper, we present a novel tracking framework with additive spatio-temporal condition, termed \textbf{ACTrack}. As shown in \cref{fig:short-c}, instead of training from scratch or fully fine-tuning the pre-trained tracker, ACTrack preserves the quality and capabilities of the pre-trained Transformer foundation model by freezing its parameters, and adds a lightweight conditional net to model the spatial correlation and temporal variations in tracking. 
Due to the patch-level global attention computation in Transformer architecture, directly capturing spatial local patterns and structures across multiple patches is tough. 
To solve this limitation, we make an additive trainable conditional net of lightweight siamese convolutional network based on similarity matching, which aims to preserve the long-term global dependencies while attending to the local generic feature modeling.
 Furthermore, the current tracking results are influenced by the previous states, which in turn affect the subsequences. Considering this, we perform temporal sequence modeling by directly predicting the coordinate sequence of the object frame-by-frame, thereby simplify the tracking pipeline by avoiding customized heads and post-processings. Experiments demonstrate our ACTrack method is effective, which can reduce overall training time by n × or more and likewise reduce memory consumption. It also achieves new state-of-the-art performance on several tracking benchmarks. Our main contributions are summarized as follows.

\begin{itemize}
    \item We propose ACTrack, a novel tracking framework with additive spatio-temporal conditions. By freezing the pre-trained Transformer foundation model and training an additive lightweight conditional net, the ACTrack could balance training efficiency and tracking performance.

    \item In order to take the spatio-temporal relationship as conditions, we design an additive siamese convolutional network and perform temporal sequence modeling by tracking the coordinate sequence of the object.

    \item The proposed method sets a new state-of-the-art performance on four tracking benchmarks, including VOT2020 \cite{kristan2020eighth}, LaSOT \cite{fan2019lasot}, TrackingNet \cite{muller2018trackingnet}, and GOT-10k \cite{huang2019got}.
\end{itemize}


\section{Related Work}
\label{sec:formatting}

~~\textbf{Visual object tracking.} The existing  mainstream trackers \cite{bertinetto2016fully,li2018high,xu2020siamfc++,chen2020siamese,guo2020siamcar,zhu2018distractor,li2019siamrpn++,kristan2023first}  commonly depend on template and search images matching. Most of them are based on siamese CNN backbone, and the central element in their design is the integration module responsible for feature fusion. Recently, Transformers \cite{vaswani2017attention} are introduced to the tracking field. Some works \cite{yan2021learning,chen2021transformer,wang2021transformer,fu2021stmtrack,gao2022aiatrack} based on CNN for generic and consistent features extraction and applies attention operations in high-level representation space. Other unified frameworks \cite{chen2022backbone,cui2022mixformer,ye2022joint,wei2023autoregressive} joint feature learning and relation modeling though Transformer to simplify the tracking process. However, the training efficiency and model performance of the trackers still require further improvement. Instead, we propose a novel tracking framework with additive spatio-temporal
conditions named ACTrack. 
By freezing the parameters of pre-trained Transformer backbone, we make an additive lightweight siamese convolutional net to preserve
the long-range global dependencies while attending to the local generic features modeling. Our tracker could balance training efficiency and performance.


\textbf{Additive learning.} To overcome the issue of large-scale training parameters, additive learning addresses it by preserving the original model weights while introducing a limited number of new parameters through techniques such as learned weight masks \cite{mallya2018piggyback,rosenfeld2018incremental}, pruning \cite{mallya2018packnet}, or hard attention \cite{serra2018overcoming}. Furthermore, Side-tuning \cite{zhang2020side,sung2022lst} incorporates a side branch model to learn additional functionality by linearly combining the outputs of a frozen model and an added networks. ControlNet \cite{zhang2023adding} locks the productionready large diffusion models, and add spatial conditioning controls to facilitate wider applications. Inspired by the above, we introduced additive learning into our ACTrack method. We freeze the parameters of Transformer foundation model and add a lightweight conditional net to reduce training memory requirements by more substantial amounts.

\begin{figure*}
  \centering
  \includegraphics[width=1\textwidth]{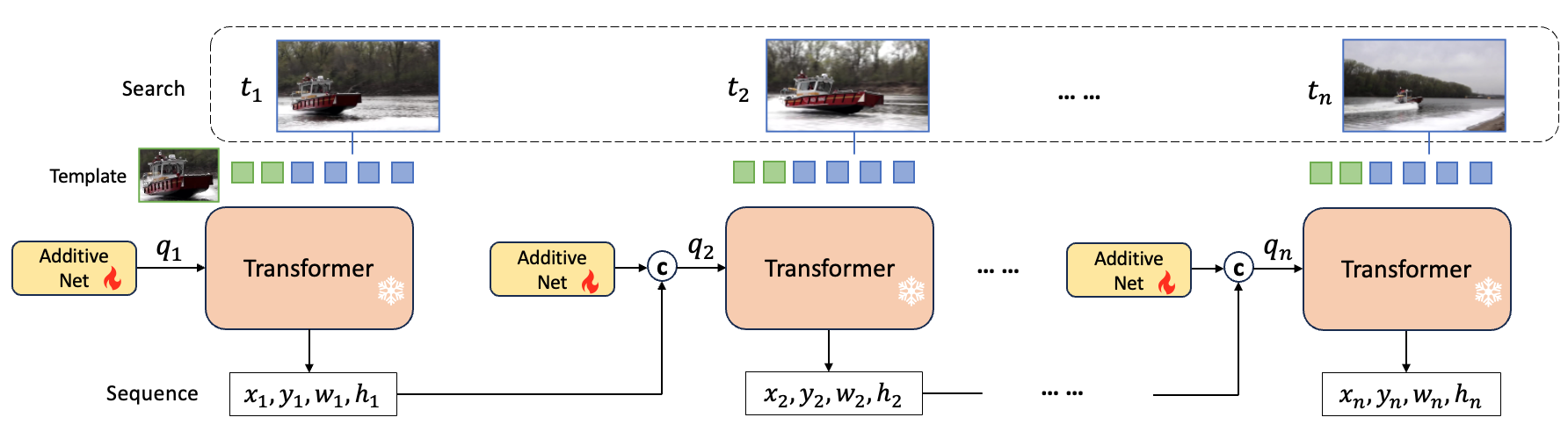}
  
  \caption{The overall architecture of \textbf{ACTrack}. The parameters of pre-trained Transformer backbone are frozen, and trainable additive 
net extracts generic features of template and search images. The track sequence are transferred from previous frame and concatenated with generic features to generate spatio-temporal conditional queries. The tracking object  coordinate is predicted frame-by-frame through temporal sequence modeling.}
  \label{fig2}
\end{figure*}

\textbf{Sequence modeling.}
Sequence learning is originally proposed for natural language modeling \cite{cho2014learning,sutskever2014sequence}. Recently, some methods apply sequence learning to computer vision, aiming to create a joint representation model for language and vision tasks \cite{bai2021connecting,li2019visualbert}.
Pix2Seq \cite{chen2021pix2seq} is a representative work that casts object detection as a token generation task conditioned on the observed pixel inputs. It represents bounding boxes and class labels as discrete sequences. Inspired by Pix2Seq, several works \cite{chen2023seqtrack,wei2023autoregressive} propose sequence modeling frameworks for visual object tracking, eliminating unnecessary task-specific heads and post-processing. In this paper, we introduce the idea of sequence modeling into our ACTrack, constructing a temporal sequence model for direct object coordinate sequence tracking frame-by-frame.


\section{Method}
See \cref{fig2}.

{
    \small
    \bibliographystyle{ieeenat_fullname}
    \bibliography{main}
}


\end{document}